%% file: emnlp2021.tex
\pdfoutput=1
\documentclass[11pt]{article}

\usepackage{emnlp2021}
\usepackage{times}
\usepackage{latexsym}

\usepackage[T1]{fontenc}
\usepackage[utf8]{inputenc}

\usepackage{microtype}

\usepackage{hyperref}
\usepackage{times}
\usepackage{latexsym}
\usepackage{booktabs}
\usepackage{tabularx}
\usepackage{todonotes}
\usepackage{multirow}
\usepackage{url}
\usepackage{amsmath}
\usepackage{mathtools}
\usepackage{amssymb}
\usepackage{amsthm}
\usepackage{tablefootnote}
\usepackage{caption}
\usepackage{subcaption}
\usepackage{bbm}
\usepackage{comment}
\usepackage{xspace}
\usepackage{algorithm}
\usepackage{proof}
\usepackage{stmaryrd}
\usepackage{bm}
\usepackage{pifont}
\usepackage{float}

\usepackage{xcolor}
\definecolor{bgblue}{rgb}{0.867,0.925,0.984}

\newcommand{\diffmask}{\textsc{DiffMask}}
\newcommand{\intgrad}{\textsc{IntGrad}}
\newcommand{\lime}{\textsc{Lime}}
\newcommand{\shap}{\textsc{Shap}}
\newcommand{\attnattr}{\textsc{AtAttr}}
\newcommand{\laattnnattr}{\textsc{LAtAttr}}
\newcommand{\arch}{\textsc{Archip}}
\newcommand{\roberta}{\textsc{RoBERTa}}
\newcommand{\annot}[1]{\colorbox{bgblue}{#1}}

\newcommand{\squad}{\textsc{Squad}}
\newcommand{\sqadv}{\textsc{Squad-Adv}}
\newcommand{\hotpot}{\textsc{HotpotQA}}

\title{Connecting Attributions and QA Model Behavior\\on Realistic Counterfactuals}

\author{Xi Ye\quad\quad Rohan Nair\quad\quad Greg Durrett\\
  Department of Computer Science \\
  The University of Texas at Austin \\
  \texttt{\{xiye,rnair,gdurrett\}@cs.utexas.edu} \\ }

\begin{document}
\maketitle
\begin{abstract}

When a model attribution technique highlights a particular part of the input, a user might understand this highlight as making a statement about counterfactuals \cite{millersurvey}: if that part of the input were to change, the model's prediction might change as well. This paper investigates how well different attribution techniques align with this assumption on \emph{realistic} counterfactuals in the case of reading comprehension (RC). RC is a particularly challenging test case, as token-level attributions that have been extensively studied in other NLP tasks such as sentiment analysis are less suitable to represent the reasoning that RC models perform. We construct counterfactual sets for three different RC settings, and through heuristics that can connect attribution methods' outputs to high-level model behavior, we can evaluate how useful different attribution methods and even different formats are for understanding counterfactuals. We find that pairwise attributions are better suited to RC than token-level attributions across these different RC settings, with our best performance coming from a modification that we propose to an existing pairwise attribution method.\footnote{Code and data:  \href{https://github.com/xiye17/EvalQAExpl}{https://github.com/xiye17/EvalQAExpl}}
\end{abstract}

\begin{figure}[t]
\centering
\includegraphics[width=\linewidth,trim=350 120 350 120,clip]{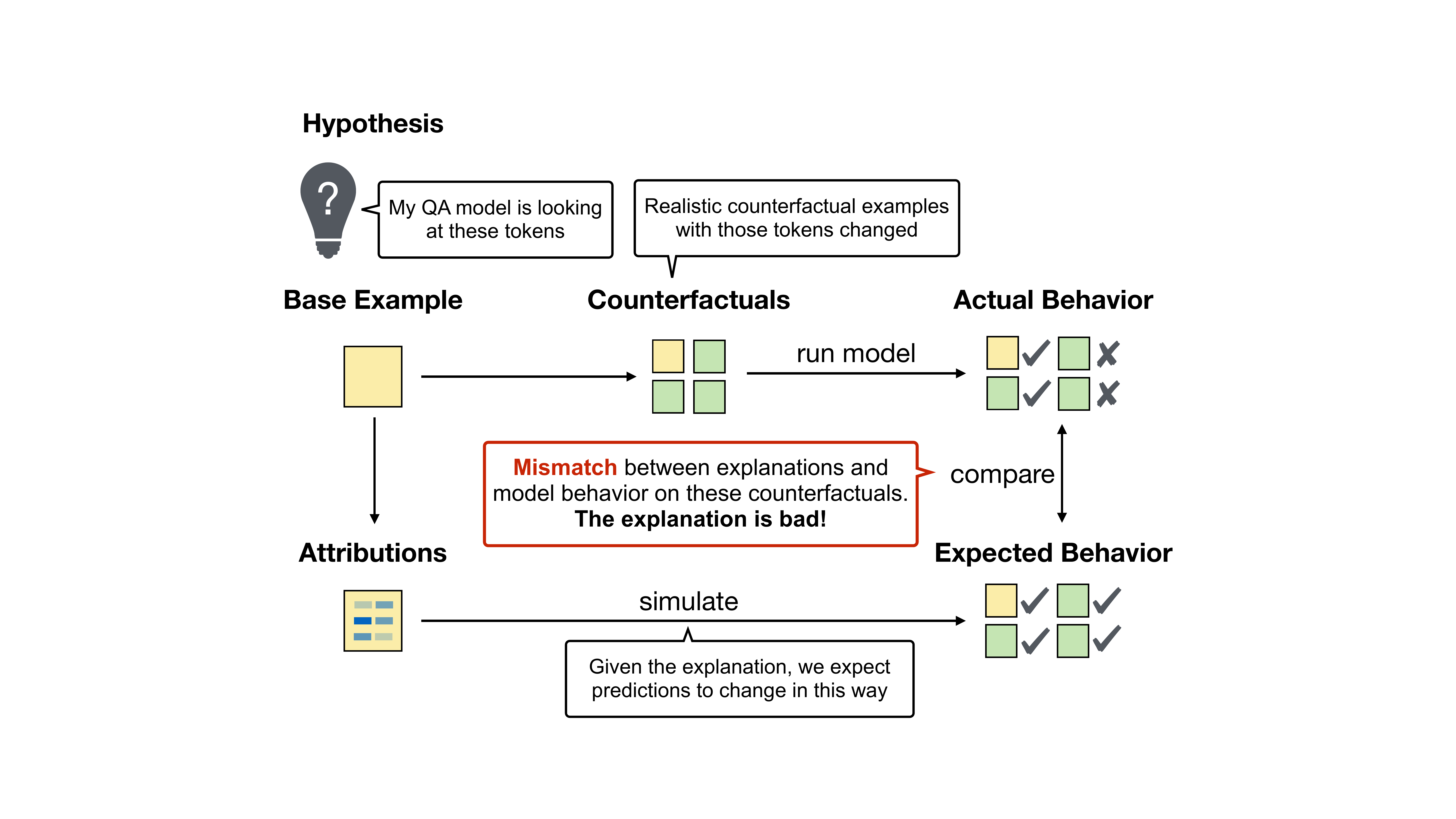}
\caption{Our methodology. Given a base example, we can formulate a hypothesis about the model's behavior, like a theory about how the model is using certain tokens. Next, we collect counterfactual examples that modify these tokens and profile the actual model behavior. Finally, we assess whether feature attributions suggest behavior consistent with what we observe, verifying whether our attributions actually enable meaningful statements about behavior on counterfactuals. }
%\revise{maybe this caption needs revising}}
\label{fig:intro}
\end{figure}

\section{Introduction}

Recent research in interpretability of neural models \cite{mythos} has yielded numerous post-hoc explanation methods, including token attribution techniques \cite{lime, intgrad, nlpexplainer,diffmask}. Attributions are a flexible explanation format and can be applied to many domains, including sentiment analysis \cite{nlpexplainer, diffmask}, visual recognition \cite{simonyan2013}, and natural language inference \cite{Camburu2018esnli,thorne-etal-2019-generating}. However, it is hard to evaluate whether these explanations are \emph{faithful} to the computation of the original model \cite{Wu2018FaithfulME,evalai, Wiegreffe2020MeasuringAB,jacovi2020towards} and as a result, they can potentially mislead users \cite{rudin2019stop}. Furthermore, attributions do not have a consistent and meaningful \emph{social attribution} \cite{millersurvey,goldberg}: that is, when a user of the system looks at an explanation, they do not necessarily draw a valid conclusion from it, making it hard to use for downstream tasks. 

How can we evaluate whether these attributions make faithful and meaningful statements about model behavior?
% Our focus on this work is how to evaluate explanations for reading comprehension in terms of their ability to reveal the \textbf{high-level} behavior of models.
In this work, we show how to use counterfactual examples to evaluate attributions' ability to reveal the \textbf{high-level} behavior of models.
%Our focus in this work is to investigate whether explanations for reading comprehension are capable of conveying the \textbf{high-level} behavior of models.
That is, rather than a vague statement like ``this word was important,'' we want attributions to give concrete, testable conclusions like ``the model compared these two words to reach its decision;'' this statement can be evaluated for faithfulness and it helps a user make important inferences about how the system behaves.
We approach this evaluation from a perspective of simulatability \cite{evalai}: can we predict how the system will behave on new or modified examples? Doing so is particularly challenging for the RC models we focus on in this work due to the complex nature of the task, which fundamentally involves a correspondence between a question and a supporting text context.

Figure~\ref{fig:intro} shows our methodology. Our approach requires annotating small sets of 
%Our core technique is to assess how well various explanations can support or reject hypotheses about the model's behavior (i.e., simulate the model) on
realistic counterfactuals, which are perturbations of original data points. These resemble several prior ``stress tests'' used to evaluate models, including counterfactual sets \cite{Kaushik2020Learning}, contrast sets \cite{contrastset}, and checklists \cite{checklist}. We first semi-automatically curate these sets to answer questions like: if different facts were shown in the context, how would the model behave? If different amounts of text or other incorrect paragraphs were retrieved by an upstream retrieval system, would the model still get the right answer?

We run the model on counterfactuals to assess the ground truth behavior. Then, given attributions from various techniques, can we predict how the model would behave \textbf{based purely on these explanations}? Our approach to do this is specific to each dataset and attribution method, but generally involves assessing how strongly the attribution method highlights tokens that are counterfactually altered, which would indicate that those tokens should impact the prediction if changed.

To showcase the kind of evaluation this method can enable, we investigate two paradigms of explanation techniques: token attribution-based \cite{simonyan2013, lime, diffmask} and feature interaction-based \cite{arch,attrattn}, which attribute decisions to sets of tokens or pairwise token interactions. For both techniques, we devise methods to connect these explanations to our high-level hypotheses about behavior on counterfactual examples. On two types of questions from \hotpot{} \cite{hotpot} and questions from adversarial \squad{} \cite{squaddataset}, we show that token-level attribution is not sufficient for analyzing RC models, which naturally involves more complex reasoning over multiple clues. We further propose a modification to an existing interaction technique from \citet{attrattn} and show improved performance on our datasets.

\begin{figure*}[t]
\centering
\includegraphics[width=\textwidth,trim=10 170 10 170,clip]{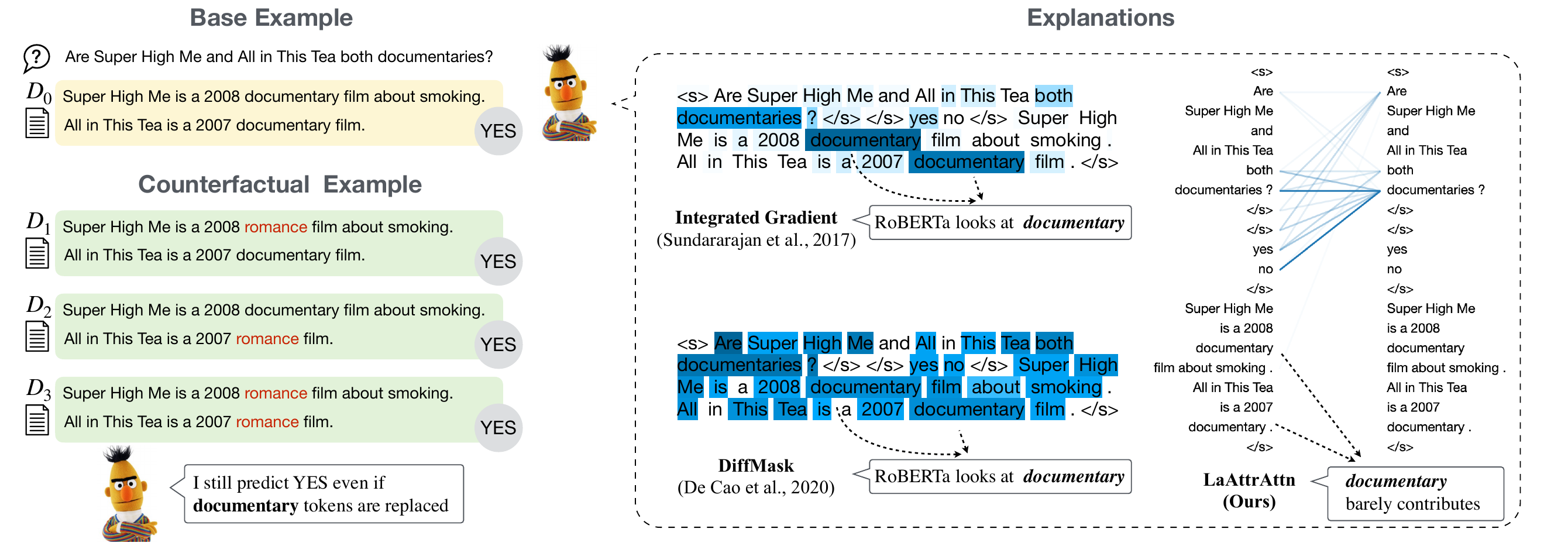}
\caption{A motivating example and attributions generated by three methods. We profile the model's behavior with its predictions on realistic counterfactual inputs, which suggest the model does not truly base its prediction on the two movies being documentaries. We can evaluate attributions by heuristically assessing whether the attribution mass yields the same conclusion about model behavior.}
%\revise{maybe this caption needs revising}}
\label{fig:motivating}
\end{figure*}

% This enables comparing different formats of explanations in a unified way.

% We apply our methodology to automatically evaluate and compare a series of explanation techniques on two types of questions from \hotpot{} \cite{hotpot}, questions from adversarial \squad{} \cite{squaddataset}, and on a synthetic QA setting.

%Our experimental results show moderate success of this approach overall, and that explanations in form of feature interactions better align with model behaviours. 

%More specifically, we analyzed \intgrad{}, \diffmask{}, \attnattr{}, archipelago. Besides, we propose a effective fix for atten. Our fixed version better align with the social attribution on real-datasets (Squad, and hotpot), compared to others. Our analysis highlights, exposes the issues with qualitatively evaluation. Our further discussion sheds light on the possibility of practical application of explanations beyond merely understanding the models.

% We summarize our main contributions as follows: (1) We propose a framework for evaluating explanations based on model simulation on realistic counterfactuals. (2) We describe a technique for connecting low-level attributions (token-level or higher-order) with high-level model hypotheses. (3) We improve an attention-based pairwise attribution technique with a simple but effective fix, leading to strong empirical results. (4) We analyze a set of QA tasks and show that our approach can derive meaningful conclusions on each.

Our main contributions are: (1) We propose a new goal for attributions, namely automatically simulating model behavior on realistic counterfactuals. (2) We describe a technique for connecting low-level attributions (token-level or higher-order) with high-level model hypotheses. (3) We improve an attention-based pairwise attribution technique with a simple but effective fix, leading to strong empirical results. (4) We analyze a set of QA tasks and show that our approach can derive meaningful conclusions about counterfactuals on each. Overall, we establish a methodology for analyzing explanations that we believe can be adapted to studying attribution methods on a wide range of other tasks with appropriate counterfactuals.

\section{Motivation}
\label{sec:motivating}

% We start by going through a detailed example of how to use our methodology to compare several attribution techniques.
We start with an example of how model attributions can be used to understand model behavior and consequently how to use our methodology to compare different attribution techniques.
Figure~\ref{fig:motivating} shows an example of a multi-hop yes/no question from HotpotQA. The QA model correctly answers \emph{yes} in this case. Given the original example, the explanations produced using \intgrad{} \cite{intgrad} and \diffmask{} \cite{diffmask} (explained in Section~\ref{sec:expl_techniques}) both assign high attribution scores to the two \emph{documentary} tokens appearing in the context: a user of the system is likely to impute that the model is comparing these two values, as it's natural to assume this model is using the highlighted information correctly. By contrast, the pairwise attribution approach we propose in this work (Section~\ref{sec:latattr}) attributes the prediction to interactions with the question, suggesting the interactions related to \emph{documentary} do not matter.

We manually curate a set of contrastive examples to test this hypothesis. If the model truly recognizes that both movies are documentaries, then replacing either or both of the \emph{documentary} tokens with \emph{romance} should change the prediction. To verify that, we perturb the original example to obtain another three examples (left side of Figure~\ref{fig:motivating}). These four examples together form a local neighborhood \cite{lime, Kaushik2020Learning, contrastset} consisting of realistic counterfactuals.\footnote{One could argue that these counterfactuals are not entirely realistic: a romance film about smoking is unlikely. Generating suitable counterfactuals is a very hard problem \cite{lianhui}, requiring deep world knowledge of what scenarios make sense or what properties hold for certain entities. The ``true'' set of realistic counterfactuals is highly domain-specific, but nevertheless, a good explanation technique should work well on a range of counterfactuals like those considered here.}
Unlike what's suggested by the token attribution based techniques, the model always predicts ``yes'' for every example in the neighborhood, casting doubt on whether the model is following the right reasoning process. Although the pairwise attribution seemed at first glance much less \emph{plausible} than that generated by the other techniques, it was actually better from the perspective of \emph{faithfully} simulating the model's behavior on these examples.% The rest of this paper will tackle two problems: first, how to formalize this process of connecting low-level feature attributions and high-level behavior, and second, how to evaluate and ``close the loop'' on understanding this pipeline on RC problems.

Our main assumption in this work can be stated as follows: \textbf{an explanation should describe model behavior with respect to realistic counterfactuals.} Past work has evaluated along plausibility criteria \cite{lei-etal-2016-rationalizing,strout-etal-2019-human,thorne-etal-2019-generating}, but as we see from this example, faithful explanations \cite{faithfulnmn,jacovi2020towards, goldberg} are better aligned with our goal of simulatability. We argue that a good explanation is one that aligns with the model's high-level behavior, from which we can understand how the model generalizes to new data. How to interpret behavior from explanations is still an open question, but we take initial steps in this work with techniques based on assessing the attribution ``mass'' on perturbed tokens.

\paragraph{Discussion: Realistic Counterfactuals}
Many counterfactual modifications are possible: past work has looked at injecting non-meaningful triggers \cite{advuniversal}, deleting chunks of content \cite{lime}, or evaluating interpolated input points as in \intgrad{}, all of which violate assumptions about the input distribution. In RC, masking part of the question often makes it nonsensical and we may not have strong expectations about our model's behavior in this case.\footnote{The exception is in adversarial settings; however, many adversarial attacks do not draw on real-world threat models \cite{athalye18a}, so we consider these less important.} Focusing on realistic counterfactuals, by contrast, illuminates fundamental problems with our RC models' reasoning capabilities \cite{squadadv, designchoice, oodhotpot, advhotpot}. This is the same motivation as that behind contrast sets \cite{contrastset}, but our work focuses on benchmarking explanations, not models themselves.

\section{Behavior on Counterfactuals}
\label{sec:eval_protocol}

We seek to formalize the reasoning we undertook in Figure~\ref{fig:motivating}. Using the model's explanation on a base data point, can we predict the model's behavior on the perturbed instances of that point?

\paragraph{Definitions} Given an original example $D_0$ (e.g., the top example in Figure~{\ref{fig:motivating}}), we construct a set of perturbations $\{D_1,...,D_k\}$ (e.g., the three counterfactual examples in Figure~{\ref{fig:motivating}}), which together with $D_0$ form a local neighborhood $\mathcal{D}$. These perturbations are realistic inputs derived from existing datasets or which we construct.

We formulate a hypothesis $\mathcal{H}$ about the neighborhood. In Figure~\ref{fig:motivating}, $\mathcal{H}$ is ``the model is comparing the target properties'' (\emph{documentary} in this case). Based on the model's behavior on the set $\mathcal{D}$, we can derive a high-level behavioral label $z$ corresponding to the truth of $\mathcal{H}$. We form our local neighborhood to check the answer empirically and compute a ground truth for $z$. Since the model always predicts ``yes'' in this neighborhood, we label set $\mathcal{D}$ with $z=0$ (the model is not comparing the properties). We label $\mathcal{D}$ as $z=1$, when the model does predict ``no'' for some perturbations.

\paragraph{Procedure} Our approach is as follows:

1. Formulate a hypothesis $\mathcal{H}$ about the model

2. Collect realistic counterfactuals $\mathcal{D}$ to test $\mathcal{H}$ empirically for some base examples

3. Use the explanation of each base example to predict $z$. That is, learn the mapping $D_0 \rightarrow z$ based on the explanation of $D_0$ so we can \textbf{simulate the model} on $\mathcal{D}$ without observing the perturbations.

Note that this third step \emph{only} uses the explanation of the \emph{base} data point: explanations should let us make conclusions about new counterfactuals without having to do inference on them.

\paragraph{Simulation from attributions} In our experiments on \hotpot{} and \squad{}, we compute a scalar factor $f$ for each attribution representing the importance of a specific part of the inputs (e.g., the \emph{documentary} tokens in Figure~\ref{fig:motivating}), which we believe should correlate with model predictions on the counterfactuals. If an attribution assigns higher importance to this information, it suggests that the model will actually change its behavior on these new examples.

Given this factor, we construct a simple classifier where we predict $z= 1$ if the factor $f$ is above a threshold.  We expect the factors extracted using better attribution methods should better indicate the model behavior. Hence, we evaluate the explanation using the \textbf{best simulation accuracy it can achieve} and the AUC score (S-ACC and S-AUC).\footnote{We do not collect large enough datasets to train a simulation model, but given larger collections of counterfactuals, this is another approach one could take.}
% Then, we are able evaluate the capability of the factor in indicating the model high-level behavior using the best simulation accuracy it can achieve and the AUC score.  That is, we construct a classification task using the factors as the features, with the goal of determining how well the factors can indicate the model high-level behavior.

Our evaluation resembles the human evaluation in \citet{evalai}, which asks human raters to predict a model's decision given an example together with its explanations, addressing simulatability from a user-focused angle. Our method differs in that (1) we automatically extract a factor to predict model behavior instead of asking humans to do so, and (2) we predict the behavior on unseen counterfactuals given the explanation of a single base data point.
% \paragraph{Evaluation} Our core evaluation metric that we use is \textbf{model simulatability}.
% using the model's explanation on a base data point, can we predict the model's behavior on the perturbed instances of that point?

\section{Explanation Techniques}
\label{sec:expl_techniques}

Compared to classification tasks like sentiment analysis, RC more fundamentally involves interaction between input features, especially between a question and a context. This work will directly compare feature interaction explanations with token attribution techniques that are more common for other tasks.\footnote{A potentially even more powerful format would be a program approximating the model's behavior, as has been explored in the context of reinforcement learning \cite{verma18a, bastani2018verifiable}. However, beyond limited versions of this \cite{anchor}, prior work does not show how to effectively build this type of explanation for QA at this time.}

For RC, each instance $D = (q, c, a)$, a tuple containing a question, context, and answer respectively. In the techniques we consider, $q$ and $c$ are concatenated and fed into a pre-trained transformer model, so our attribution techniques will explain predictions using both of these.

%Following our philosophy, we study feature interaction as as well as token level attribution (for comparison) as explanations for model Bert-based QA models, since it demonstrates strong improvements over traditional RNN-based models.

\subsection{Token Attribution-Based}

These techniques all return scores $s_i$ for each token $i$ in both the question and context.

\textbf{LIME} \cite{lime} and \textbf{SHAP} \cite{lundberg2017unified} both compute the attribution values for individual input features by using a linear model to locally approximate the model's predictions on a set of perturbed instances around the base data point. The attribution value for an individual input feature is the corresponding weight of the linear model. \lime{} and \shap{} are different in the way of specifying instance weights used to train the linear model: \lime{} computes the weights heuristically, whereas \shap{} uses a procedure based on Shapley values.

 \textbf{Integrated Gradient (\intgrad{})} \cite{intgrad} computes an attribution for each token by integrating the gradients of the prediction with respect to the token embeddings over the path from a baseline input (typically mask or pad tokens) towards the designated input. Although a common technique, recent work has raised concern about the effectiveness of \intgrad{} methods for NLP tasks, as interpolated word embeddings do not correspond to real input values \cite{harbecke2021explaining,sanyal2021discretized}.

\textbf{Differentiable Mask (\diffmask{})} \cite{diffmask} learns to mask out a subsets of the input tokens for a given example while maintaining a distribution over answers as close to the original distribution as possible. This mask is learned in a differentiable fashion, then a a shallow neural model (a linear layer) is trained to recognize which tokens to discard. %\diffmask{} softly masks out the input tokens with a gate, keeping the masking operation differentiable, so as to train a shallow neural model (a linear layer) to recognize which tokens to discard. 
% \diffmask{} trains to  to predic,zt using a shallow neural model

\subsection{Feature Interaction-Based}
\label{sec:inter}

These techniques all return scores $s_i$ for each pair of tokens $(i,j)$ in both the question and context that are fed into the QA system.

\textbf{Archipelago} \citep{arch} measures non-additive feature interaction. Similar to \diffmask{}, \arch{} is also implicitly based on unrealistic counterfactuals which remove tokens. Given a subset of tokens, \arch{} defines the contribution of the interaction by the prediction obtained from masking out all the other tokens, only leaving a very small fraction of the input. %E.g., the interaction between ``not bad'' in the ``a not bad movie'', is defined as the prediction score obtained with the input utterance ``[mask] not bad [mask]'' subtracting the score obtained with ``[mask] [mask] [mask] [mask]''.
Applying this definition to a complex task like QA can result in a nonsensical input.

\textbf{Attention Attribution (\attnattr{})} \cite{attrattn}
uses attention specifically to derive pairwise explanations. However, it avoids the pitfalls of directly inspecting attention \cite{attnsucks1,attnsucks2} by running an integrated gradients-like procedure over all the attention links within transformers, yielding attribution scores for each link. The attribution scores directly reflect the attribution of the particular attention links, making this model able to describe pairwise interactions.

Concretely, define the $h$-head attention matrix over input $D$ with $n$ tokens as $A=[A_1,...,A_l]$, where $A_i\in \mathbb{R}^{h\times n \times n}$ is the attention scores for each layer. We can obtain the attribution score for each entry in the attention matrix $A$ as:
%\scriptsize
\begin{equation}
\label{eq:attnattr}
\mathrm{ATTR}(A)=A \odot \int_{\alpha=0}^1 \frac{\partial F(D, \alpha A)}{\partial A} d\alpha,
\end{equation}
%\normalsize
where $F(D, \alpha A)$ is the transformer model that takes as input the tokens and a matrix specifying the attention scores for each layer. We later sum up the attention attributions across all heads and layers to obtain the pairwise interaction between token $(i,j)$, i.e., $s_{ij}=\sum_m\sum_n \mathrm{ATTR(A)_{mnij}}$.

\subsection{Layer-wise Attention Attribution}
\label{sec:latattr}

\begin{figure}[t]
\centering
\includegraphics[width=\linewidth,trim=150 180 110 190,clip]{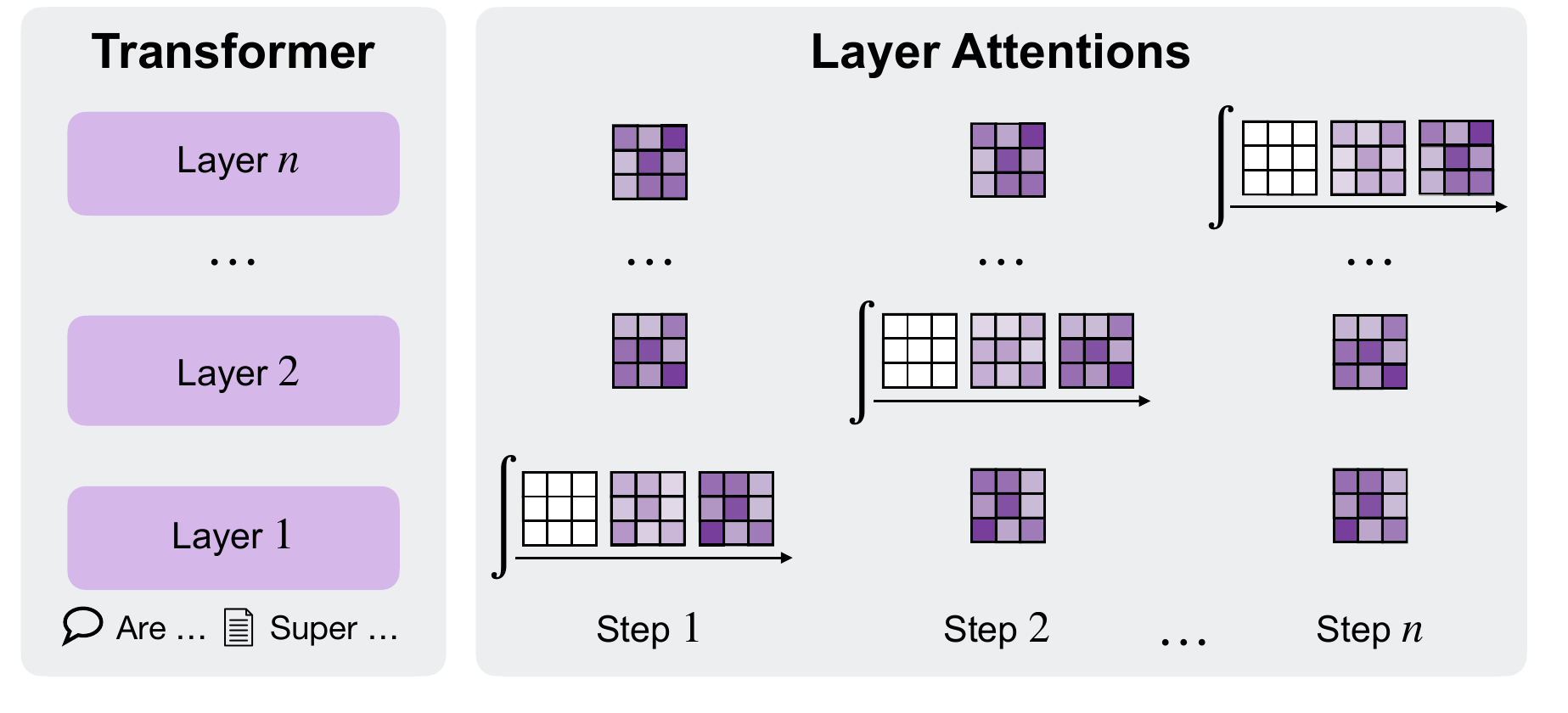}
\small
\caption{Steps of our Layer-wise Attention Attribution approach, where we modify a single layer's attention at a time. For example, to compute the attribution of attentions at layer 2, we only intervene on the attention matrix at that layer, and leave the other attentions computed as usual. }
\label{fig:layerattns}
% \vspace{-1.5em}
\end{figure}

We propose a new technique \laattnnattr{} to improve upon \attnattr{} for the RC setting. The \attnattr{} approach simultaneously increases all attention scores when computing the attribution, which could be problematic. Since the attention scores of higher layers are determined by the attention scores of lower layers, forcibly setting all the attention scores and computing gradients at the same time may distort the gradients for the lower level links and produce inaccurate attribution. When applying \intgrad{} approach in other contexts, we typically assume the independence of input features (e.g., pixels of an image and tokens of an utterance), an assumption that does not hold here.

To address this issue, we propose a simple fix, namely applying the \intgrad{} method layer-by-layer. As shown in Figure~\ref{fig:layerattns}, to compute the attribution for attention links of layer $i$, we only change the attention scores at layer $i$:
%\scriptsize
\begin{equation}
\label{eq:laattnattr}
\mathrm{ATTR}(A_i)=A_i \odot \int_{\alpha=0}^1 \frac{\partial F_{/i}(D, \alpha A_i)}{\partial A_i} d\alpha.
\end{equation}
%\normalsize
$F_{/i}(D, \alpha A_i)$ denotes that we only intervene on the attention masks at layer $i$ while leaving other attention masks computed naturally via the model. We pool to obtain the final attribution for pairwise interaction as $s_{ij}=\sum_m\sum_n \mathrm{ATTR(A)_{mnij}}$.
% This technique does not necessarily satisfy the Completeness axiom commonly used in this line of work \cite{intgrad}. Since our ultimate goal is a downstream empirical evaluation, we set aside any theoretical analysis of this technique for now.

% These perturbations are realistic constrastive cases from existing datasets or which we construct. We automatically extract a factor to connect the explanation and the model behavior. In our experiments, a factor is usually the importance of a specific part of the inputs (E.g., the ``documentary'' tokens in Figure~\ref{fig:motivating}), which correlates to model predictions on the foils. 

% We expect the factors extracted using good explanations should indicates the model behavior well. Hence, we evaluate the explanation using {\bf simulatability accuracy and AUC}. That is, we construct a classification task using the factors as the features, with the goal of determining how well the factors can indicate the model high-level behavior.

%To profile the model's high-level behavior, we will use existing adversarial contrast examples or construct new ones for two datasets. In this work, we attempt to create realistic contrastive cases. The difficulty of these cases is somewhat less relevant; we have examples that challenge models to varying degrees, since our goal is to assess whether the explanations can distinguish between examples that will confuse the model and those that won't.

\section{Experiments}

% We evaluate our attribution methods (Section~\ref{sec:expl_techniques}) follow our stated evaluation protocol (Section~\ref{sec:eval_protocol}) on the \hotpot{} dataset \cite{hotpot}, and the \squad{} dataset \cite{squaddataset}, specifically leveraging examples from adversarial \squad{} \cite{squadadv}.
We assess whether attributions can achieve our proposed goal following the setup in Section~\ref{sec:eval_protocol} on the \hotpot{} dataset \cite{hotpot}, and the \squad{} dataset \cite{squaddataset}, specifically leveraging examples from adversarial \squad{} \cite{squadadv}.

\paragraph{Implementation Details}
For experiments on \hotpot{}, we base our analysis on a \roberta{} \cite{roberta} QA model in the distractor setting. We implement our model using the Huggingface library \cite{huggingface} and train the model for 4 epochs with a learning rate of 3e-5, a batch size of 32, and a warm-up ratio of 6\%. Our model achieves 77.2 F1 on the development set in the distractor setting, comparable to other strong \roberta{}-based models \cite{sae,groeneveld2020}.

In the \sqadv{} setting, we also use a \roberta{} QA model which achieves 92.2 F1 on the \squad{} dev set and 68.0 F1 on \sqadv{}. Our model is trained on \squad{} v1.0 for 4 epochs using a learning rate of 1.5e-5, a batch size of 32 and a warm-up ratio of 6\%.

%However, past work has shown that models trained on \squad{} can be easily attacked \cite{squadadv}; models trained on \hotpot{} take significant reasoning shortcuts \cite{adv,designchoice,oodhotpot}. Therefore, explanations on this dataset would be valuable to have if they could easily identify these.

%with each question designed to require consulting two documents. 
% However, past work has shown that models take significant reasoning shortcuts \cite{adv designchoice,oodhotpot,advhotpot}; explanations on this dataset would be valuable to have if they could easily identify these.

% \begin{figure*}[t]
% \centering
% \includegraphics[width=\linewidth]{figures/squad-ex.png}
% \caption{squad example}
% \label{fig:squadexs}
% \end{figure*}

% \paragraph{Data}

\begin{figure}[t]
    \scriptsize
    \centering
    \begin{tabularx}{\linewidth}{|c  X|}
    \toprule
     \multirow{9}{*}{(a)} & {\bf \texttt{Question:}\,} Were Ulrich Walter and Léopold Eyharts both from Germany?  \\
     & {\bf \texttt{Context:}\,} Léopold Eyharts (born April 28, 1957) is a Brigadier General in the \annot{French} Air Force, an engineer and ESA astronaut. \\
     & Prof. Dr. Ulrich Hans Walter (born February 9, 1954) is a \annot{German} physicist/engineer and a former DFVLR astronaut. \\
     &{\bf \texttt{Substitutes:}\,} French, German \\
     \midrule
    \multirow{11}{*}{(b)}  &{\bf \texttt{Question:}\,} Are the movies "Monsters, Inc." and "Mary Poppins" both by the same company?   \\
       & {\bf \texttt{Context:}\,} Mary Poppins is a 1964 American musical-fantasy film directed by Robert Stevenson and produced by \annot{Walt Disney}, with songs written and composed by the Sherman Brothers. \\
       & Monsters, Inc. is a 2001 American computer-animated comedy film produced by Pixar Animation Studios and distributed by \annot{Walt Disney} Pictures. \\
     &{\bf \texttt{Substitutes:}\,} Walt Disney, Universal \\
    \midrule
    \multirow{14}{*}{(c)}  & {\bf \texttt{Question:}\,} \annot{What} was the father of Kasper Schmeichel \annot{voted to be by the IFFHS in 1992?}\\
       & {\bf \texttt{Context:}\,} Peter Bolesław Schmeichel MBE (born 18 November 1963) is a Danish former professional footballer who was voted the IFFHS World's Best Goalkeeper in 1992 and 1993. \\
       &Kasper Peter Schmeichel (born 5 November 1986) is a Danish professional footballer. He is the son of former Manchester United and Danish international goalkeeper Manuel Neuer.\\
 &{\bf \texttt{AdvSent1:}\,} Robert Lewandowski was voted to be the World's Best Striker in 1992. \\
    &{\bf \texttt{AdvSent2:}\,} Michael Jordan was voted the IFFHS best NBA player in 1992. \\
     \bottomrule
    \end{tabularx}
    \caption{Examples (contexts are truncated for brevity) of our property annotations on HotpotQA base data points. The top two are yes/no questions and the third is a bridge question.}
    % We find the property tokens in the context, and build realist counterfactuals by replacing them with substitutes that are properties extracted in the base data point or similar properties hand-selected by us.}
    \label{fig:hpqa_annotation}
\end{figure}

\subsection{Hotpot Yes-No Questions}
\label{sec:exp_yesno}
%We investigate explanations for \hotpot{}, which requires multi-hop reasoning.
% and consequently is considered a harder task than \squad{}.
% As pointed out by previous research \cite{badhotpot}, models can often leverages reasoning shortcuts in the Bridge Type Question. 
We first study a subset of yes/no comparison questions, which are challenging despite the binary answer space \cite{clark-etal-2019-boolq}.
Typically, a yes-no comparison type question requires comparing the properties of two entities (Figure~\ref{fig:motivating}). 

\paragraph{Hypothesis \& Counterfactuals}
The hypothesis $\mathcal{H}$ we investigate is as in Section~\ref{sec:motivating}: \emph{the model compares the entities' properties as indicated by the question.}
Most Hotpot Yes-No questions follow one of two templates: \emph{Are A and B both \_\_?} (Figure~\ref{fig:hpqa_annotation}a), and \emph{Are A and B of the same \_\_?} (Figure~\ref{fig:hpqa_annotation}b). We define the \textbf{property} tokens associated with each question as the tokens \emph{in the context} that match the blank in the template; that is, the values of the property that A and B are being compared on. For example, in Figure~\ref{fig:hpqa_annotation}a, \emph{French} and \emph{German} are the property tokens, as the property of interest is the national origin.

%We first mark the property tokens associated with the two entities that are being compared. For the formal template, we select the tokens indicating whether A or B is \_\_; for the latter template, we select the tokens that answer the  \_\_ of A or B.

To construct a neighborhood for a base data point, we take the following steps: (1) manually extract the property tokens in the context; (2) replace the property token with two substitutes, forming a set of four counterfactuals exhibiting nonidentical ground truths.
When the properties associated with the two entities differ from each other, we directly use the properties extracted as the substitutes (Figure~\ref{fig:hpqa_annotation}a); otherwise we add a new property candidate that is of the same class (Figure~\ref{fig:hpqa_annotation}b).

% \revise{ For instance, for the question \emph{Are A and B of the same nationality}, the properties are nationalities of ``A'' and ``B''; for the question \emph{Are A and B both ice plants}, the properties are their plant species.
% As in the motivating example, we later construct the counterfactuals by replacing the properties in the context with the one another if the two properties are different or similar hand-selected properties (e.g., ``documentary'' $\rightarrow$ ``romance'', ``American'' $\rightarrow$ ``English'') if the two are the same, producing additional three perturbations $D_1,D_2,D_3$ for each base example $D_0$.}
We set $z=0$ (the hypothesis does not hold) if for each perturbed example $D_i \in \mathcal{D}$, the model predicts the same answer as for the original example, indicating a failure to compare the properties. We set $z=1$ if the model's prediction \emph{does} change. We choose a binary scheme to label the model behavior because we observed that, on the small perturbation sets, the model performance was bimodal: either the tokens mattered for the prediction (reflected by the model changing its prediction at least once) or they didn't.
% However, for a larger perturbation set (like 10+ examples) where performance is less stable, a more nuanced notion of performance would be helpful.
 The authors annotated perturbations for 50 ($\mathcal{D}$, $z$) randomly selected pairs in total, forming a total of 200 counterfactual instances. 
Full counterfactual sets are available with our data and code release.
% More examples of the annotation process and concrete examples can be found in the Appendix.

% We now have a classification task to learn the mapping $D_0 \rightarrow z$ based on the explanation of $D_0$; that is, can we successfully predict the behavior on $\mathcal{D}$ without observing these perturbations?

\begin{table}[t]
\footnotesize
    \centering
    \begin{tabular}{lcccc}
    \toprule 
        \multirow{2}{*}{Approach}  & \multicolumn{2}{c}{Yes-No} &  \multicolumn{2}{c}{Bridge}  \\
        &  S-ACC & S-AUC & S-ACC & S-AUC \\
         \midrule
         {\sc Majority} & 52.0 & $-$& 56.0 & $-$ \\
        {\sc Conf} & 64.0 & 49.8$^\dagger$ & 66.0 & 65.9$^\dagger$ \\
        \toprule
         \lime{}  & 72.0 & 73.6$^\dagger$ & 74.0 & 71.4$^\dagger$  \\
          \shap{}  & 72.0 & 70.5$^\dagger$ & 76.0 & 75.0\phantom{$^\dagger$}  \\
         %\midrule
         \intgrad{}  & 72.0 & 75.2$^\dagger$ & 72.0 & 77.9\phantom{$^\dagger$}  \\
         \diffmask{} & 66.0 & 60.2$^\dagger$ & 68.0 & 62.3$^\dagger$\\
         \midrule
         \arch{}  & 56.0 & 53.2$^\dagger$ & 62.0 & 57.5$^\dagger$ \\
         \attnattr{} & 66.0 & 63.6$^\dagger$ & 72.0 & 79.1\phantom{$^\dagger$} \\
         \laattnnattr{}  & \bf 84.0 & \bf 87.9\phantom{$^\dagger$} & \bf 78.0 & \bf 81.7\phantom{$^\dagger$} \\
         \bottomrule
    \end{tabular}
    \caption{Results on \hotpot{} Yes-No type and Bridge questions. %Our approach can better predict the model behavior on realistic counterfactuals, surpassing token attribution methods.
    We perform significance tests on accuracy via bootstrap resampling for the comparisons between \laattnnattr{} and other approaches. A dagger indicates a method for which our approach outperforms it by a statistically significant margin ($p < 0.05$). Overall, the best pairwise technique yields better simulation accuracy than the best token-level technique.}
    \label{tab:hotpot}
\end{table}

\paragraph{Connecting Explanation and Hypothesis} %Given a large number of examples, we could theoretically learn a model to predict the behavior based on the explanations. In this work, we instead
To make a judgment about $z$, we extract a factor $f$ based on the importance of a set of property tokens $P$. For token attribution-based methods, we define $f$ as the sum of the attribution $s_i$ of each token in $P$: $\sum_{i\in P} s_i$. For feature interaction-based methods producing pairwise attribution $s_{ij}$, we compute $f$ by pooling the scores of all the interaction related to the property tokens: $\sum_{i\in P \vee  j \in P } s_{ij}$.

Now we predict $z=1$ if the factor $f$ is above a threshold, and evaluate the capability of the factor in indicating the model high-level behavior using the best simulation accuracy it can achieve (S-ACC) and AUC score (S-AUC).\footnote{Note that for different attribution methods, the thresholds are different and set to achieve the best accuracy.}

\paragraph{Results}
 % Similarly to the \squad{} setting, we measure how well different techniques align with model behaviours using the correlation between the importance of the properties indicated by the explanations and the label of if the model relies on the properties. We show the results in Table~\ref{tab:hotpot}.
 
%   \diffmask{} and our \laattnnattr{} are more effective compared to other techniques for this task, hitting a optimal accuracy of around 80\%. That means, with a proper set threshold, we can successfully predict whether a model is robust w.r.t. to an input example 80\% of the times. Besides, our approach still outperforms vanilla \attnattr{}.
First, we show that using attributions can indeed help predict the model's behavior. In Table~\ref{tab:hotpot}, our approach (\laattnnattr{}) is the best, achieving a simulation accuracy of 84\%. That is, with a properly set threshold, we can successfully predict whether the model predictions change when perturbing the properties in the original example 84\% of the time. The attributions therefore give us the ability to simulate our model's behavior better than the other methods here. Our approach also improves substantially over the vanilla \attnattr{} method.

% Token attribution based approaches obtain an accuracy around 72\%. This indicates token attribution based methods are not effective in the \hotpot{} setting which engages with interaction between tokens more intensively.
The best token-level attribution based approaches obtain an accuracy of 72\%, significantly lagging the best pairwise technique. This indicates token attribution based methods are less effective in the \hotpot{} Yes-No setting; we hypothesize that this is due to the importance of token interaction in this RC setting.

In this setting, \diffmask{} performs poorly, typically because it assigns high attribution to many tokens, since it determines which tokens need to be kept rather than distinguishing fine-grained importance (see the appendix for examples). It's possible that other heuristics or models learned on large numbers of perturbations could more meaningfully extract predictions from this technique.

\begin{figure}[t]
\centering
\includegraphics[width=0.8\linewidth,trim=650 250 650 250,clip]{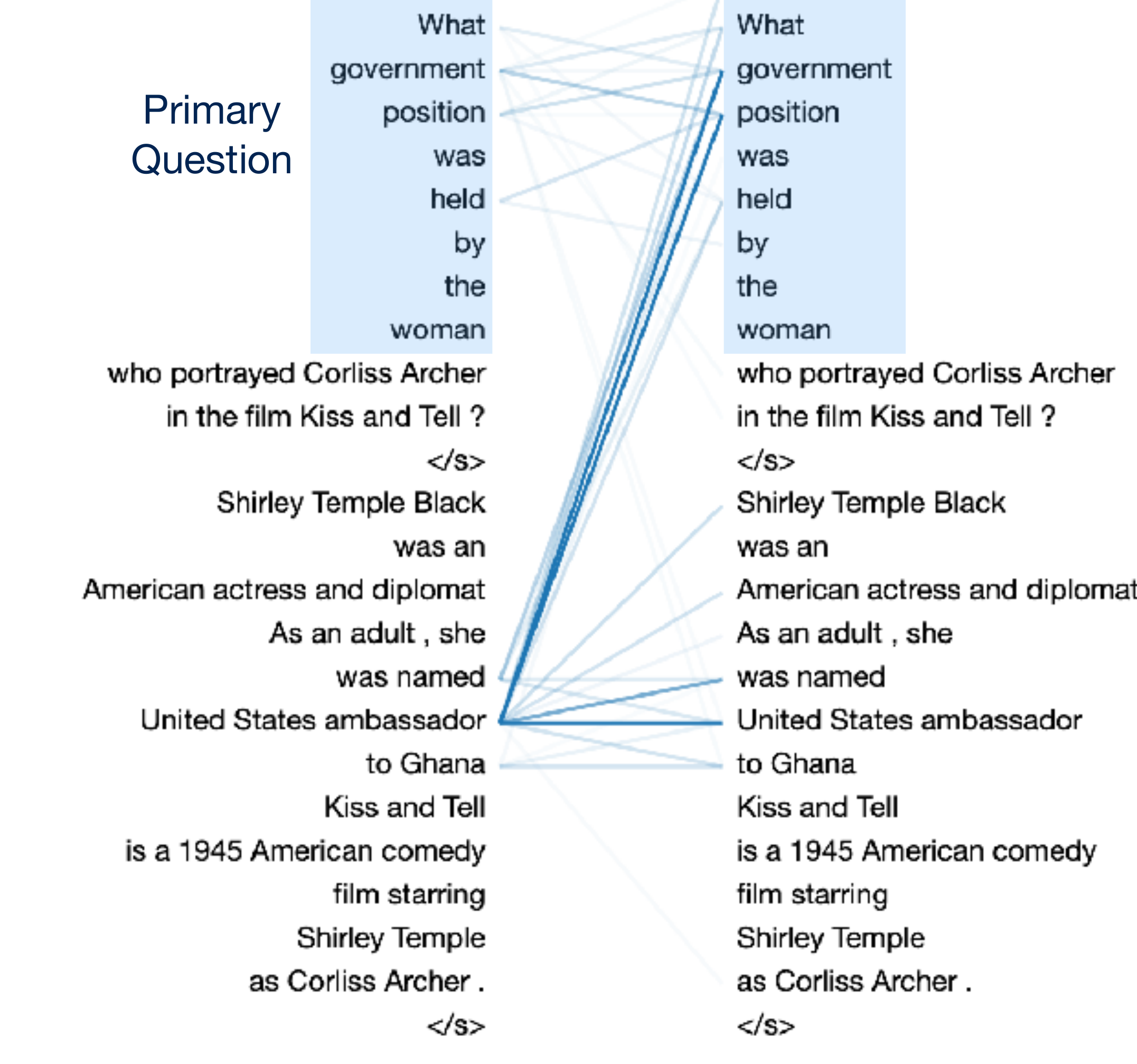}
\caption{Explanations generated by our approach for a bridge question from \hotpot{}. The prediction can mostly be attributed to the primary question, indicating the model is taking a reasoning shortcut, and the prediction is flipped with an adversarial attack. }
\label{fig:bridgeexs}
\end{figure}

\subsection{Hotpot Bridge Questions}
We also evaluate the explanation approaches on so-called bridge questions on the \hotpot{} dataset, described in \citet{hotpot}. Figure~\ref{fig:bridgeexs} shows an example explanation of a bridge question. From the attribution scores we find the most salient connection is between the span ``\emph{what government position}'' in the question and the span ``\emph{United States Ambassador}'' in the context. This attribution directly highlights the reasoning shortcut \cite{squadadv, designchoice, oodhotpot, advhotpot} the model is using, where it disregards the second part of the question.
 % Our approach exposes how the model skips the reasoning steps by directly highlighting the shortcuts.
If we inject an additional sentence \textit{ ``Hillary Clinton is an American politician, who served as the United States secretary of the state from 2009 to 2013''} into the context, the model will be misled and predict ``\emph{United States secretary}'' as the new answer. This sentence could easily have been part of another document retrieved in the retrieval stage, so we consider its inclusion to be a realistic counterfactual.

We further define the \emph{primary} question, i.e., the span of the question containing wh-words with heavy modifier phrases and embedded clauses dropped, following the decomposition principle from \citet{oodhotpot}. In Figure~\ref{fig:bridgeexs}), the primary question is ``\emph{What government position is held by the woman}.''

\paragraph{Hypothesis \& Counterfactuals} The hypothesis $\mathcal{H}$ we investigate is: \emph{the model is using correct reasoning and not a shortcut driven by the primary question}.

We construct counterfactuals following the same idea applied in our example. We view bridge questions as consisting of two single hop questions, the primary part and the secondary part. 
For a given question, we add an adversarial sentence based on the primary question so as to alter the model prediction.
The added adversarial sentence contains context leading to a spurious answer to only the primary question, but does not change the gold answer.
We do this twice, yielding a set $\mathcal{D} = \{D_0, D_1, D_2\}$ consisting of the base example and two perturbations. We define the label of $D$ to be $z=0$ in the case that model's prediction does change under the perturbations, and $z=1$ otherwise. We show one example in Figure~\ref{fig:hpqa_annotation}c. More examples and the full counterfactual set can be found in the appendix.
% Again, we try to predict based on the explanation of $D_0$ whether the model's behavior will remain constant over this set or change.

We randomly sample 50 base data points from the development set and two authors each write an adversarial sentence, giving 150 data points total.

\paragraph{Connecting Explanation and Hypothesis}
For this setting, we use a factor describing the importance of the primary question normalized by the importance of the entire question. Namely, let $P=\{p_i\}$ be the set of tokens in the primary questions, and $Q=\{q_i\}$ be the set of tokens in the entire question. We define the factor $f$ as the importance of $P$ normalized by the importance of $Q$, where the importance calculation is the same as in Section~\ref{sec:exp_yesno}.
A higher factor means it is more heavily relying only on the primary question and hence a better chance of being attacked.

\paragraph{Results}

According to the simulation ACC scores in Table~\ref{tab:hotpot}, token-level attributions are somewhat more successful at indicating model behavior in this setting compared to the yes/no setting. Our approach as the best feature interaction based technique is able to achieve a stimulation accuracy of 78\%, slightly outperforming the best token attribution approach.

\subsection{SQuAD Adversarial}

%Modern QA models have already achieved super-human performance on \squad{}, but can be easily fooled with \cite{squadadv}, since the model usually relies no superficial spurious correlation to pick the answer instead of following the desired reasoning process. For this setting, we base our analysis on a \roberta{} model achieving 92.1 F1 scores on the development set.

\paragraph{Hypothesis \& Counterfactuals} Our hypothesis $\mathcal{H}$ is: \emph{the model can resist adversarial attacks of the addSent variety} \cite{squadadv}. For each of the original examples $D_0$ from a portion of the \sqadv{} development set, \citet{squadadv} creates 5 adversarial attacks, which are paraphrased and filtered by Turkers to give 0 to 5 valid attacks for each example, yielding our set $\mathcal{D}$. We define the label of $\mathcal{D}$ to be $z=1$ if the model resists all the adversarial attacks posed on $D_0$ (i.e., predictions for $D$ are the same). To ensure the behavior is more precisely profiled by the counterfactuals, we only keep the base examples with more than 3 valid attacks, resulting in a total number of 276 $(\mathcal{D},z)$ pair (1,506 data points).

\paragraph{Connecting Explanation and Hypothesis}
We use a factor $f$ indicating the importance of the essential keywords extracted from the question using POS tags (proper nouns and numbers). E.g., for the question ``\emph{What Florida stadium was considered for Super Bowl 50}'', we extract ``\emph{Florida}'', ``\emph{Super Bowl}'' , and ``\emph{50}''. If the model considers all the essential keywords mentioned in the question, it should not be fooled by distractors with irrelevant information. We show a set of illustrative examples in the appendix. We compute the importance scores in the same way described in Section~\ref{sec:exp_yesno}.
 % In the first example, the model heavily considers ABC and New York when making the prediction, suggesting it truly realizes the informations needed for finding the answer. Therefore, it model will more likely to be robust when adversarial senteces are injected. In contrast, the model seems to ignore the entities in the second example, so it is likely to be more vulnerable to adversarial attack.

% This is based on the underlying  More specifically, for each example we extract the value for variable $x$, the attribution of a set of essential keywords, and observe the label for variable $y$, based on if the model resists all adversarial attacks, and we assess how well $x$ indicates $y$. In other word, we can image the a binary classification task where the importance of the keywords are the only feature. Since entities mentioned in the question are generally important,  we automatically extract the keywords by applying an NER system on the questions.
In addition to the scores provided by various explanation techniques, we also use the model's confidence on the original prediction as a baseline.
% \begin{figure}[t]
% \centering
% AUC scores

% of different explanation techniques.
% \label{fig:squadauc}
% \end{figure}

\begin{table}[t!]
\small
    \centering
    \begin{tabular}{lcc}
    \toprule
        Approach & S-ACC & S-AUC   \\
        \midrule
        \textsc{Majority} & 52.1 & $-$  \\
        \textsc{Conf} & 58.3 & 57.8$^\dagger$ \\
        \midrule
         \lime{} & 67.7 & 68.3$^\dagger$ \\
         \shap{} & 65.9 & 68.3$^\dagger$ \\
        \midrule
         \intgrad{} & 61.6 & 61.1$^\dagger$ \\
         \diffmask{} & 57.6 & 53.6$^\dagger$ \\
         \midrule
         \arch{} &  58.6 & 56.2$^\dagger$ \\
         \attnattr{} &  69.4 & \bf 72.5\phantom{$^\dagger$}\\
         \laattnnattr{} & \bf 70.0 & 72.1\phantom{$^\dagger$}\\ 
         \bottomrule
    \end{tabular}
    \caption{Simulation Accuracy and AUC scores for the SQuAD adversarial setting, assessing whether the model changes its prediction on an example when attacked. We perform significance tests on accuracy via bootstrap resampling for the comparisons between \laattnnattr{} and other approaches. A dagger indicates a method for which our approach outperforms it by a statistically significant margin ($p < 0.05$).} %Dagger indicates a method for which our approach outperforms it by a statistically significant margin.
    \label{tab:squad}
\end{table}

% \begin{table}[h]
% \footnotesize
%     \centering
%     \begin{tabular}{l|c|c|c|c|c|c}
%      &   CO & {Int-} & {Diff-} &{Archi-} & {Attr-} & {LaAttr-} \\
%   ACC & NF & {Grad}  & {Mask} & {pelago} &  {Attn} & {Attn} \\
%          \bottomrule
%     \end{tabular}
%     \caption{Simulation Accuracy and AUC scores for assessing whether model changes its prediction on an example when being attacked.}
%     \label{tab:squad}
% \end{table}
\paragraph{Results} We show results in Table~\ref{tab:squad}. The best approaches (\attnattr{} and \laattnnattr{}) can achieve a simulation accuracy around 70\%, 10\% above the performance based on raw model confidence. This shows the model is indeed over-confident in its predictions; our assumption about the robustness together with our technique can successfully expose the vulnerability in some of the model predictions.

There is room to improve on these results; our simple heuristic cannot perfectly connect the explanations to the model behavior in all cases. We note that there are other orthogonal approaches \cite{kamath-etal-2020-selective} to calibrate the confidence of QA models' predictions by looking at statistics of the adversarial examples. Because our goal is to assess attributions rather than optimize for calibration, our judgment is made purely based on the \emph{original} example, and does not exploit learning to refine our heuristic.

% \end{comment}
\subsection{Discussion and Limitations}

% \paragraph{Post-hoc Calibration using Explanation}
% As shown in previous section, the explanations can help to assess if the model is robust under certain input examples. We can also use this feature for calibration. To the best of our knowledge, we are the first to demonstrate the practical applications of explanations techniques beyond understanding.

% \paragraph{Failures of Human Evaluation}
% A core argument we are push is human evaluation is not sufficient because a human-certified good explanation might not align with model behaviours under realistic foils. We have shown an real example from \hotpot{} in Section~\ref{sec:motivating}. here we present an additional example from xxx.

We show that feature attributions can reveal known dataset biases and reasoning shortcuts in HotpotQA without having to perform a detailed manual analysis. This confirms the suitability of our attribution methods for at least this use case: model designers can look at them in a semi-automated way and determine how robust the model is going to be when faced with counterfactuals.

% \paragraph{Qualitatively Comparison between Token Attribution and Feature Interaction Based}
% For bridge type, we manually inspect some examples, interaction based techniques can tell you what kind of shortcuts the model is leveraging, wheraa
Our analysis also highlights the limitations of current explanation techniques.
We experimented with other counterfactuals by permuting the order of the paragraphs in the context, which often gave rise to different predictions. We believe the model prediction was in these cases impacted by biases in positional embeddings (e.g., the answer tends to occur in the first retrieved paragraph), which cannot be indicated by current attribution methods. We believe this is a useful avenue for future investigation. By first thinking about what kind of counterfactuals and what kind of behaviours we want to explain, we can motivate the development of new explanation techniques to serve these needs.

\section{Related Work}

We focus on several prominent token attribution techniques, but there are other related methods as well, including other methods based on Shapley values \cite{vstrumbelj2014explaining, lundberg2017unified}, contextual decomposition \cite{Jin2020Towards}, and hierarchical explanations \citep{chenGenerating2020}. These formats can also be evaluated using our framework if being connected with model behavior using the proper heuristic.
Other work explores so-called concept-based explanations \cite{mu2020compositional, bau2017network, Yeh2019OnCE}. These provide another pathway towards building explanations of high-level behavior; however, they have been explored primarily for image recognition tasks and cannot be directly applied to QA, where defining these sorts of concepts is challenging. Finally, textual explanations \cite{Hendricks2016GeneratingVE} are another popular alternative, but it is difficult to evaluate these in our framework as it is very difficult to bridge from a free-text explanation to an approximation of a model's computation.
% \paragraph{Probing}

Probing techniques aim to discover what intermediate representations have been learned in neural models \cite{tenney2018what,conneau-etal-2018-cram,hewitt2019control,voitainformation}. Internal representations could potentially be used to predict behavior on contrast sets similar to this work; however, this cannot be done heuristically and larger datasets are needed to explore this.

% \paragraph{Evaluation of Explanation} 
Other work considering how to evaluate explanations
% by requiring humans to predict model decisions given explanations\cite{chandrasekaran2018,nguyencomparing,evalai}, 
is primarily based on how explanations can assist humans in predicting model decisions for a given example \cite{doshi2017towards,chandrasekaran2018,nguyencomparing,evalai};
% but mostly focusing on plausibility \cite{}.
We are the first to consider building contrast sets for this. Similar ideas have been used in other contexts \cite{Kaushik2020Learning, contrastset} but our work focuses on evaluation of explanations rather than general model evaluation.

\section{Conclusion}
% We have presented an evaluation technique based on realistic counterfactuals to evaluate explanations for RC models. We show that our evaluation method distinguishes which explanations truly give us insight about high-level model behavior. 
We have presented a new methodology using explanations to understand model behavior on realistic counterfactuals. We show explanations can indeed be connected to model behavior, and therefore we can compare explanations to understand which ones truly give us actionable insights about what our models are doing.
% Feature interaction-based techniques perform the best in our analysis, especially our \laattnnattr{} method; we believe that this could be a useful evaluation paradigm if extended and formalized across a range of tasks.

We have showcased how to apply our methodology on several RC tasks, leveraging either semi-automatically curated counterfactual sets or existing resources. We generally find pairwise interaction methods perform better than the best token-level attribution based methods in our analysis. More broadly, we see our methodology as a useful evaluation paradigm that could be extended across a range of tasks, leveraging either existing contrast sets or with a small amount of effort devoted to create targeted counterfactual sets as in this work. 

\section*{Acknowledgments}

Thanks to Eunsol Choi, Jiacheng Xu, Jifan Chen, Qiaochu Chen, and everyone in the UT TAUR lab for helpful discussions, as well as to the anonymous reviewers for their helpful feedback.
This work was partially supported by NSF Grant IIS-1814522, a gift from Arm, a gift from Salesforce Inc, and an equipment grant from NVIDIA.

\bibliography{anthology,custom}
\bibliographystyle{acl_natbib}

\appendix
\input{appendix}

\end{document}

%% file: appendix.tex
\onecolumn
\newpage
\twocolumn

\section{Details of Hotpot Yes-No Counterfactuals}
\begin{figure*}[t]
    \centering
    \small
    \begin{tabularx}{\linewidth}{|c  c  X|}
    \toprule
     \multirow{6}{*}{(a)} & Question & Were Ulrich Walter and Léopold Eyharts both from Germany?  \\
     & Context &Léopold Eyharts (born April 28, 1957) is a Brigadier General in the \annot{French} Air Force, an engineer and ESA astronaut. \\
     && Prof. Dr. Ulrich Hans Walter (born February 9, 1954) is a \annot{German} physicist/engineer and a former DFVLR astronaut. \\
     &Substitutes & French, German \\
     \midrule
    \multirow{5}{*}{(b)} &Question & Are both Aloinopsis and Eriogonum ice plants? \\
     &Context & Aloinopsis is a genus of \annot{ice} plants from South Africa. \\
     && Eriogonum is the scientific name for a genus of \annot{flowering} plants in the family   Polygonaceae. The genus is found in North America and is known as wild buckwheat. \\
    &Substitutes & ice, flowering \\
     \midrule
    \multirow{6}{*}{(c)} &Question & Were Frank R. Strayer and Krzysztof Kieślowski both Directors? \\
     &Context & Frank R. Strayer (September 21, 1891 - 2013 February 3, 1964) was an actor, film writer, and \annot{director}. He was active from the mid-1920s until the early 1950s. \\
    && Krzysztof Kieślowski (27 June 1941 - 13 March 1996) was a Polish art-house film \annot{director} and screenwriter. \\
    &Substitutes & director, producer \\
      \midrule
      \midrule
      \multirow{5}{*}{(d)}  &Question & Were Scott Derrickson and Ed Wood of the same nationality? \\
        &Context & Scott Derrickson (born July 16, 1966) is an \annot{American} director, screenwriter and producer. \\
        && Edward Davis Wood Jr. (October 10, 1924 - 2013 December 10, 1978) was an \annot{American} filmmaker, actor, writer, producer, and director. \\
    &Substitutes & American, English \\
        \midrule
     \multirow{6}{*}{(e)}  &Question & Are the movies "Monsters, Inc." and "Mary Poppins" both by the same company?   \\
       &Context & Mary Poppins is a 1964 American musical-fantasy film directed by Robert Stevenson and produced by \annot{Walt Disney}, with songs written and composed by the Sherman Brothers. \\
       && Monsters, Inc. is a 2001 American computer-animated comedy film produced by Pixar Animation Studios and distributed by \annot{Walt Disney} Pictures. \\
     &Substitutes & Walt Disney, Universal \\
       \midrule
    \multirow{6}{*}{(f)}  & Question & Are Steve Perry and Dennis Lyxzén both members of the same band?\\
       &Context & Stephen Ray Perry (born January 22, 1949) is an American singer, songwriter and record producer. He is best known as the lead singer of the rock band \annot{Journey}. \\
      && Dennis Lyxzén (born June 19, 1972) is a musician best known as the lead vocalist for Swedish hardcore punk band \annot{Refused}. \\
    &Substitutes &Journey, Refused \\
        % \midrule
        % Q: Are TEC-1 and Dubna 48K are based on the same processor?\\
        % The Dubna 48K is a Soviet clone of the ZX Spectrum home computer. It was based on an analogue of the Zilog Z80 microprocessor. \\
        % The TEC-1 is a single-board kit computer first produced by the Australian hobbyist electronics magazine Talking Electronics in the early 1980s. It was based on the Zilog Z80 CPU, had 2K of RAM and 2K of ROM in a default configuration. \\
     \bottomrule
    \end{tabularx}
    \caption{Examples (contexts are truncated for brevity) of our annotations on Hotpot Yes-No base data points. We find the property tokens in the context, and build realist counterfactuals by replacing them with substitutes that are properties extracted in the base data point or similar properties hand-selected by us.}
    \label{fig:yesno_annotation}
\end{figure*}
Figure~\ref{fig:yesno_annotation} shows several more examples to illustrate our process of generating counterfactuals for the Hotpot Yes-No setting.

As stated in Section~\ref{sec:exp_yesno}, most Hotpot Yes-No questions follow one of two templates: \emph{Are A and B both \_\_?} (Figure~\ref{fig:yesno_annotation}, abc), and \emph{Are A and B of the same \_\_?} (Figure~\ref{fig:yesno_annotation}, def). The property tokens that match the blank in the template are highlighted in Figure~\ref{fig:yesno_annotation}.

% We define the \textbf{property} tokens associated with each question as the tokens \emph{in the context} that match the blank in the template; that is, the values of the property that A and B are being compared on.
% For example, in Figure~\ref{fig:yesno_annotation}a, \emph{French} and \emph{German} are the property tokens, as the property of interest is the national origin.

%We first mark the property tokens associated with the two entities that are being compared. For the formal template, we select the tokens indicating whether A or B is \_\_; for the latter template, we select the tokens that answer the  \_\_ of A or B.

% To construct a neighborhood for a base data point, we take the following steps:
Recall our two steps to construct a neighborhood for a base data point: 
\begin{enumerate}
    \item Manually extract the property tokens in the context
    \item Replace each property token with a substitute, forming a set of four counterfactuals exhibiting nonidentical ground truths
\end{enumerate}
When the properties associated with the two entities differ from each other, we directly use the properties extracted as the substitutes (Figure~\ref{fig:yesno_annotation}, abf); otherwise we add a new property candidate that is of the same class (Figure~\ref{fig:yesno_annotation}, cde).

We annotated randomly sampled examples from the Hotpot Yes-No questions. We skipped several examples that compared abstract concepts with no explicit property tokens. For instance, we skipped the question \emph{Are both Yangzhou and Jiangyan District considered coastal cities?} whose given context does not explicitly mention whether the cities are coastal cities.
We looked through 61 examples in total and obtained annotations for 50 examples, so such discarded examples constitute a relatively small fraction of the dataset. Overall, this resulted in 200 counterfactual instances. We found the prediction of a {\sc Roberta} QA model on 52\% of the base data points change when being perturbed.

%\todo{idk if I like this after writing it. Xi: the argument is actually the same as footnote 1? maybe let's not include this}\greg{Note that the examples could be somewhat ``inconsistent'' after applying this process: for example, \emph{Ulrich Hans Walter} is more likely to be a German than a French name. However, what we are testing is whether the model's attribution to the property token accurately reflects how much this token truly matters when the data is perturbed. Our view is that these data instances are natural enough, and constitute well-formed inputs that can help us answer this question with respect to a realistic counterfactual scenario.}
% \onecolumn

\section{Details of Hotpot Bridge Counterfactuals}
Figure~\ref{fig:bridge_annotation} shows more examples of our annotations for generating counterfactuals for Hotpot Bridge examples.
% We first highlight the primary question, and later write down two adversarial sentences to attack the model.
We first decompose the bridge questions into two single hop questions \cite{oodhotpot}, the \textbf{primary} part (marked in Figure~\ref{fig:bridge_annotation}) and \textbf{secondary} part. The primary part is the main body of the question, whereas the secondary part is usually a clause used to link the bridge entity \cite{oodhotpot}.

Next, we write adversarial sentences for confuse the model follow a similar method used for generating {\sc SQuAD} adversarial examples \cite{squadadv}. Specifically, we only look at the primary part, and write down a sentence that can answer the primary question accordingly with a different entity from the secondary question. This will introduce a spurious answer, but does not change the gold answer. Besides, we also write the sentences follow in the same Wikipedia style as the original context possible, and some of the sentences are modified from texts from Wikipedia (e.g., Figure~\ref{fig:bridge_annotation} ac).

% Specifically, we view bridge questions as consisting of two single hop questions, the \textbf{primary} part (marked in Figure~\ref{fig:bridge_annotation}) and \textbf{secondary} part. The primary part is the main body of the question, whereas the secondary part is usually a clause used to link the bridge entity \cite{oodhotpot}.

% We construct our neighborhoods as follows:
% \begin{enumerate}
%     \item Manually decompose our sentences into the primary and secondary parts
%     \item Make up adversarial sentences to confuse the model. For a given base example, an adversarial sentence typically provides a spurious answer to the primary question, but does not change the gold answer.
% \end{enumerate}
%Next, we make up adversarial sentences to confuse the model. 
% In order to match the distribution of the contexts in Hotpot dataset, most of our adversarial sentences are modified from excerpts of Wikipedia articles.

Two of the authors each wrote a single adversarial sentence for 50 of the Hotpot Bridge examples, yielding 150 counterfactual instances in total. The adversarial sentences manage to alter 56\% of the predictions on the base examples.

\begin{figure*}[t]
    \small
    \centering
    \begin{tabularx}{\linewidth}{|c  c  X|}
    \toprule
     \multirow{6}{*}{(a)} & Question &\annot{What is the name of the fight song of the university} whose main campus is in Lawrence, Kansas and whose branch campuses are in the Kansas City metropolitan area?  \\
     &Context & Kansas Song (We're From Kansas) is a fight song of the University of Kansas. \\
     && The University of Kansas, often referred to as KU or Kansas, is a public research university in the U.S. state of Kansas. The main campus in Lawrence, one of the largest college towns in Kansas, is on Mount Oread, the highest elevation in Lawrence. Two branch campuses are in the Kansas City metropolitan area.  \\
     &Adv Sent 1 & Texas Fight is a fight song of the University of Texas at Austin. \\
     &Adv Sent 2 & Big C is a fight song of the University of California, Berkeley. \\
     \midrule
    \multirow{5}{*}{(b)} &Question  &  \annot{What screenwriter with credits for "Evolution"} co-wrote a film starring Nicolas Cage and Téa Leoni? \\
     &Context & David Weissman is a screenwriter and director. His film credits include "The Family Man" (2000), "Evolution" (2001), and "When in Rome" (2010). \\
     &&  The Family Man is a 2000 American romantic comedy-drama film directed by Brett Ratner, written by David Diamond and David Weissman, and starring Nicolas Cage and  Téa Leoni. \\
      &Adv Sent 1 & Don Jakoby is an American screenwriter that collabrates with David Weissman in "Evolution". \\
     &Adv Sent 2 & Damien Chazelle is a screenwriter most notably known for writing La La Land. \\
     \midrule
    \multirow{6}{*}{(c)} &Question & \annot{The arena }where the Lewiston Maineiacs played their home games \annot{ can seat how many people}? \\
    &Context  & The Androscoggin Bank Colisée (formerly Central Maine Civic Center and Lewiston Colisee) is a 4,000 capacity (3,677 seated) multi-purpose arena, in Lewiston, Maine, that opened in 1958. \\
    && The Lewiston Maineiacs were a junior ice hockey team of the Quebec Major Junior Hockey League based in Lewiston, Maine. The team played its home games at the Androscoggin Bank Colisée.  \\
   &Adv Sent 1 &Allianz (known as  Fußball Arena München for UEFA competitions) is a arena in Munich, with a 5,000 seating capacity. \\
    &Adv Sent 2 &The Tacoma Dome is a multi-purpose arena (221,000 capacity, 10,000 seated) in Tacoma, Washington, United States. \\
      \midrule
      \multirow{5}{*}{(d)}  &Question & Scott Parkin has been a vocal critic of Exxonmobil and another \annot{corporation that has operations in how many countries}? \\
        &Context & Scott Parkin (born 1969, Garland, Texas is an anti-war, environmental and global justice organizer, former community college history instructor, and a founding member of the Houston Global Awareness Collective. He has been a vocal critic of the American invasion of Iraq, and of corporations such as Exxonmobil and Halliburton.\\
        &&The Halliburton Company, an American multinational corporation. One of the world's largest oil field service companies, it has operations in more than 70 countries. \\
  &Adv Sent 1 & Visa is a corporation that has operations in more than 200 countries. \\
    &Adv Sent 2 &The Ford Motor Company is an American multinational corporation with operations in more than 100 countries. \\
        \midrule
     \multirow{6}{*}{(e)}  &Question & In 1991 Euromarché was bought by a \annot{chain that operated how any hypermarkets at the end of 2016?}  \\
       &Context & Carrefour S.A. is a French multinational retailer headquartered in Boulogne Billancourt, France, in the Hauts-de-Seine Department near Paris. It is one of the largest hypermarket chains in the world (with 1,462 hypermarkets at the end of 2016). \\
       && Euromarché was a French hypermarket chain. In June 1991, the group was rebought by its rival, Carrefour, for 5,2 billion francs.  \\
   &Adv Sent 1 & Walmart Inc is a multinational retail corporation that operates a chain of hypermarkets that owns 4,700 hypermarkets within the United States at the end of 2016. \\
    &Adv Sent 2 & Trader Joe's is an American chain of grocery stores headquartered in Monrovia, California. By the end of 2016, Trader Joe's had over 503 stores nationwide in 42 states. \\
       \midrule
    \multirow{6}{*}{(f)}  & Question & \annot{What} was the father of Kasper Schmeichel \annot{voted to be by the IFFHS in 1992?}\\
       &Context & Peter Bolesław Schmeichel MBE (born 18 November 1963) is a Danish former professional footballer who played as a goalkeeper, and was voted the IFFHS World's Best Goalkeeper in 1992 and 1993. \\
       && Kasper Peter Schmeichel (born 5 November 1986) is a Danish professional footballer. He is the son of former Manchester United and Danish international goalkeeper Manuel Neuer.\\
 &Adv Sent 1 & Robert Lewandowski was voted to be the World's Best Striker in 1992. \\
    &Adv Sent 2 &Michael Jordan was voted the IFFHS best NBA player in 1992. \\

     \bottomrule
    \end{tabularx}
    \caption{Examples (contexts are truncated for brevity) of primary questions and adversarial senteces for creating Hotpot Bridge counterfactuals.}
    \label{fig:bridge_annotation}
\end{figure*}

% \section{Details of Synthetic Dataset}
% \label{sec:detailsynth}

% Our dataset is generated using templates, with 20 entities (\texttt{E0} through \texttt{E19}) and 20 relations (\texttt{R0} through \texttt{R19}). We place 3 or 4 entities in the context. We randomly inject \texttt{<mask>} tokens between entity relation pairs (we do not inject \texttt{<mask>} within any entity relation pair) to prevent model learning spurious correlation with positional embeddings.

% We create a training/validation set of 200,000/10,000 examples, respectively, and train a 2-layer 12-head transformer model for this task, achieving 100\% accuracy on the training set and over 98\% accuracy on validation set. 

% \onecolumn
% \newpage

% \section{Additional Examples of Explanations for \squad{} }
% See Figure~\ref{fig:adex_squada}, \ref{fig:adex_squadb}.

\begin{figure*}[t]
     \centering
     \begin{subfigure}[h]{0.45\textwidth}
         \centering
         \includegraphics[width=\textwidth,trim=80 500 80 500,clip]{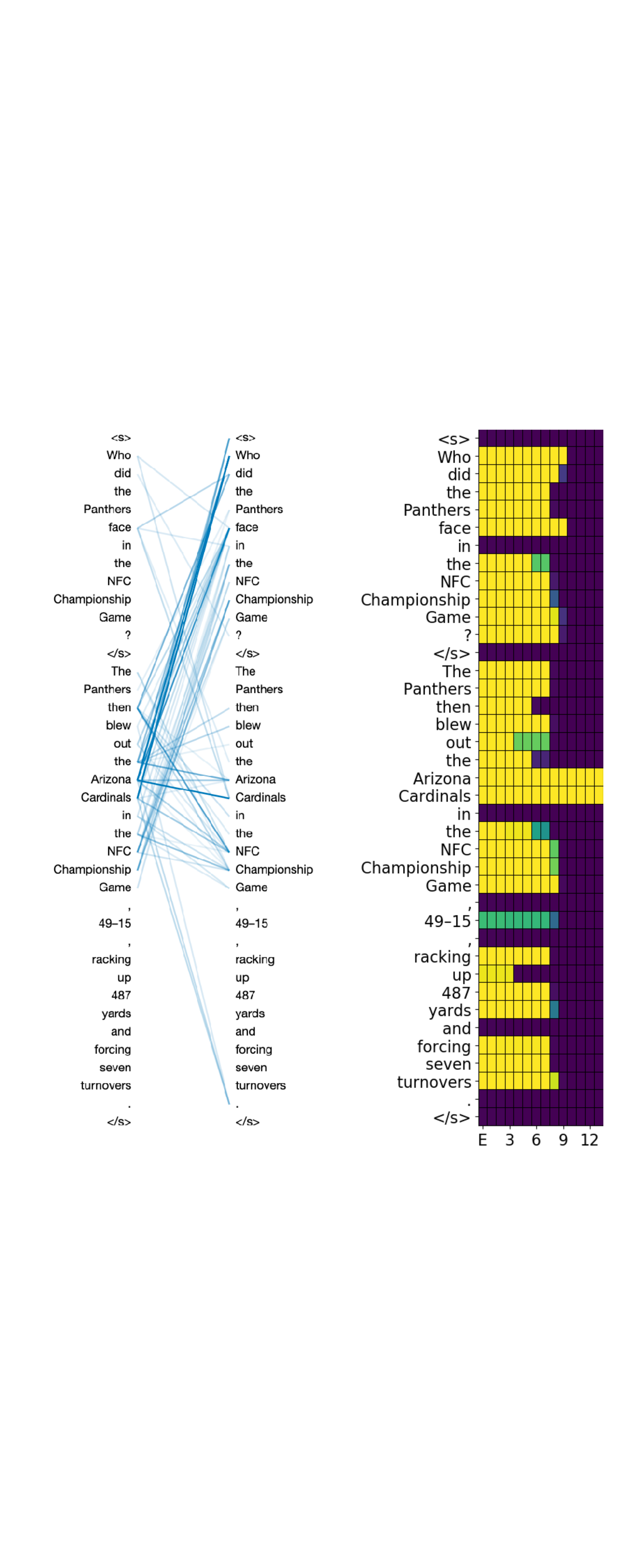}
         \caption{Explanations generated by our approach (left) and \diffmask{} (right) for an example from \squad{} dataset. We automatically extract ``Panthers'' and ``NFC Championship Game'' as the essential keywords. From the explanation, we see these keywords contribute to the model prediction, and therefore infer the model is more likely to be able to resist adversarial attacks posed on this example.}
        %  \label{fig:y equals x}
            \label{fig:adex_squada}
     \end{subfigure}
     \hfill
     \begin{subfigure}[h]{0.45\textwidth}
         \centering
         \includegraphics[width=\textwidth,trim=0 500 0 500,clip]{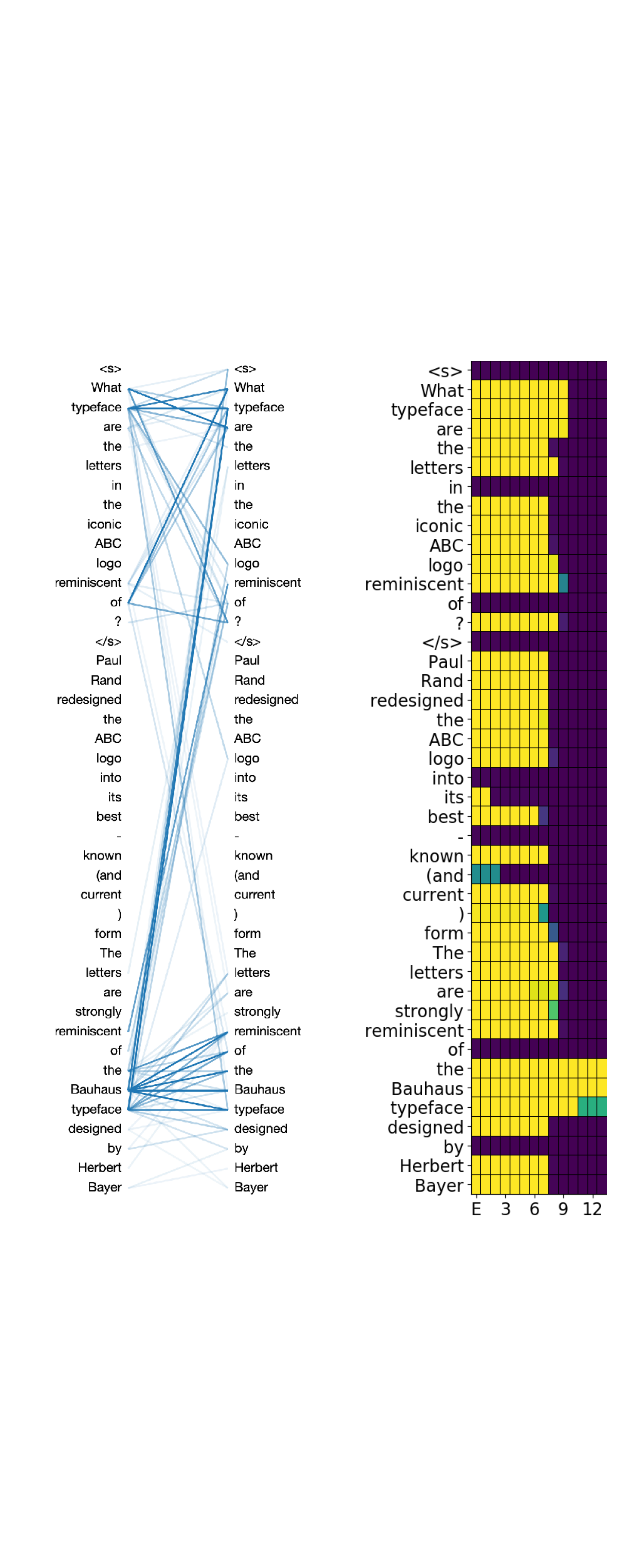}
         \caption{Explanations generated by our approach (left) and \diffmask{} (right) for an example from \squad{} dataset. When adding an adversarial sentence, the model changes its prediction. Our explanation clearly shows the model bases its prediction on ``what typeface'' without taking into account the entity ``ABC''.}
                    \label{fig:adex_squadb}

     \end{subfigure}
\end{figure*}

% \clearpage
% \newpage
% \newpage

% \section{Additional Examples of Explanations for \hotpot{} }

\begin{figure*}[t]
     \centering
     \begin{subfigure}[b]{0.45\textwidth}
         \centering
         \includegraphics[width=\textwidth]{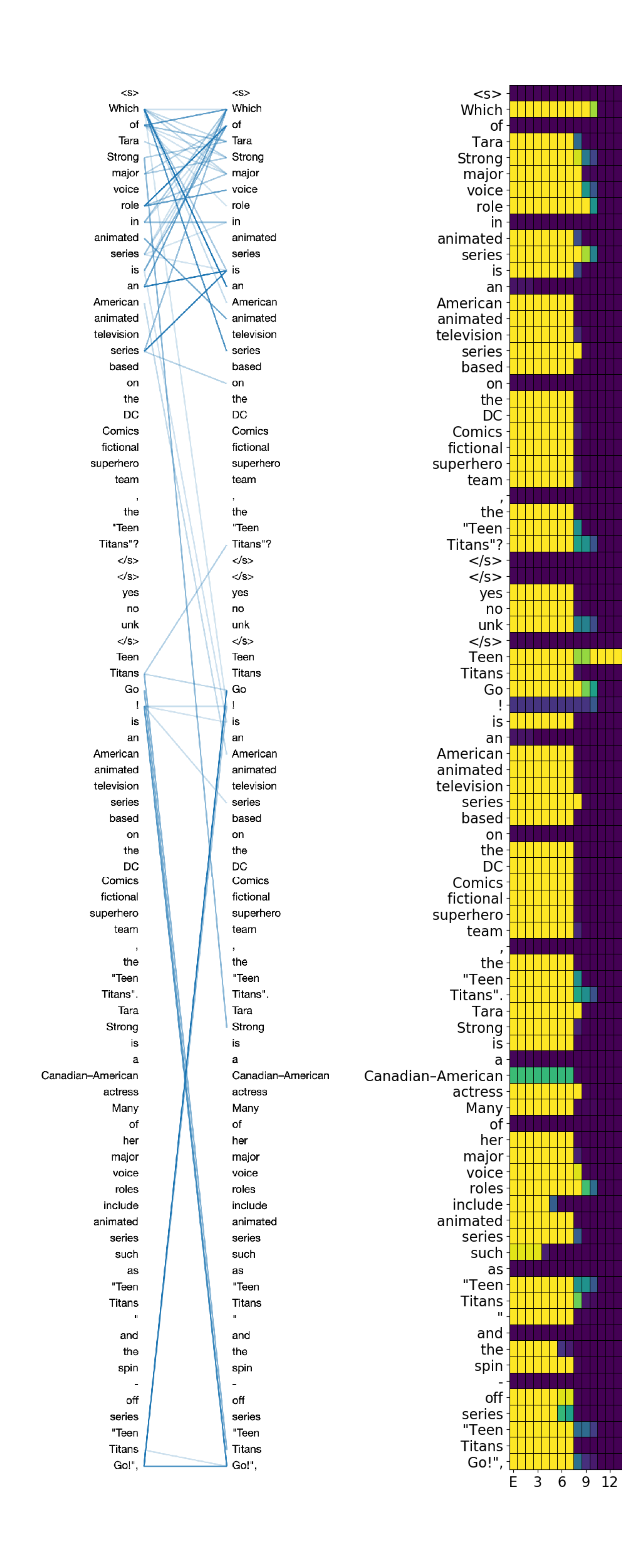}
         \caption{Explanations generated by our approach (left) and \diffmask{} (right) for a bridge example from the \hotpot{} dataset. The model can resist adversarial sentences posed on this example. Our explanation highlights certain tokens in the latter part of the question (``American animated television'' and ``Teen Titans''), suggesting the model prediction is less likely to be flipped by adversarial attacks targeted at this example, which aligns with the model behavior. }
        %  \label{fig:y equals x}
        \label{fig:adex_hotpota}
     \end{subfigure}
     \hfill
     \begin{subfigure}[b]{0.45\textwidth}
         \centering
         \includegraphics[width=\textwidth,trim=0 0 0 0,clip]{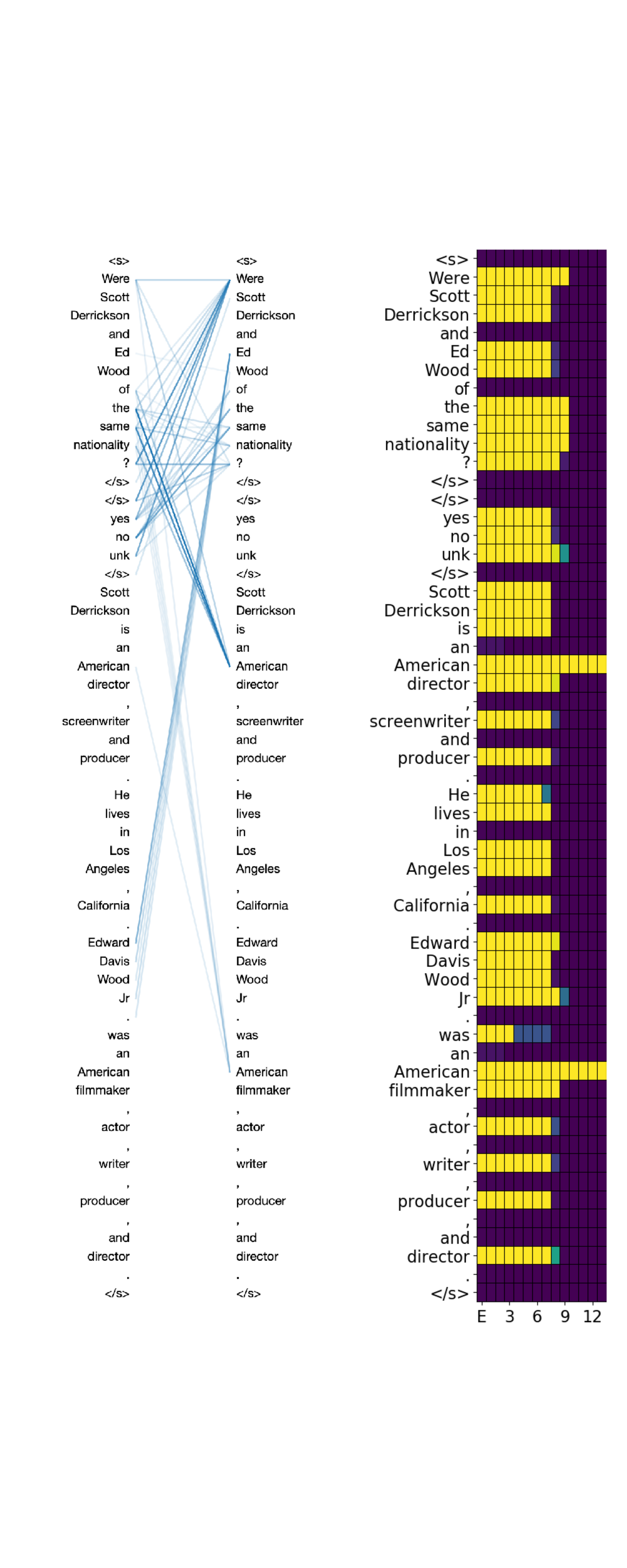}
         \caption{Explanations generated by our approach (left) and \diffmask{} (right) for a comparison example from the \hotpot{} dataset.  When we perturb the nationalities in the context, the model changes its prediction. Both of the explanations both suggest the model makes its decision by looking at the nationalities associated with the two entities, which is congruent with the model behavior.  }
        %  \label{fig:three sin x}
        \label{fig:adex_hotpotb}
     \end{subfigure}
\end{figure*}

% See Figure~\ref{fig:adex_hotpota}, ~\ref{fig:adex_hotpotb}.